\documentclass[10pt, 14paper, citep, DIV12]{scrartcl}

\usepackage{microtype}
\usepackage{graphicx}
\usepackage{xcolor}
\usepackage{subfigure}
\usepackage{booktabs} 

\usepackage{hyperref}

\usepackage{url}
\usepackage{natbib}
\usepackage{booktabs}
\usepackage{multirow}
\usepackage{graphicx}
\usepackage{amsmath,amssymb,mathtools,bm,bbm}
\usepackage{enumerate}
\usepackage{pgfplots}
\usepackage{libertine}
\usepackage{hyperref}
\usepackage{authblk}

\newcommand{\FIG}{figure}
\newcommand{\SEC}{section}

\newcommand{\fgsm}{FGSM}

\setcitestyle{authoryear,round}

\DeclareMathOperator{\sign}{\text{sign}}
\DeclareMathOperator{\clip}{\text{clip}}

\begin{document}

\title{%
\LARGE
Ensemble Methods as a Defense to Adversarial Perturbations Against 
Deep 
Neural Networks
}

\renewcommand{\thefootnote}{$\star$} 

\author{\normalsize Thilo Strauss%
\footnote{Authors contributed equally.}
\thanks{thilo.strauss@etas.com}
}
\author{\normalsize Markus Hanselmann\small*%
\thanks{markus.hanselmann@etas.com}
}
\author{\normalsize Andrej Junginger}
\author{\normalsize Holger Ulmer}

\affil{
Machine Learning Group at ETAS GmbH, Bosch Group, 
Borsigstra\ss e 14, 70469 Stuttgart, Germany
}


\publishers{%
\normalsize
}
\date{\normalsize\today}



%


\maketitle

\begin{abstract}
Deep learning has become the state of the art approach in many machine learning 
problems such as classification. 
It has recently been shown that deep learning is highly vulnerable to 
adversarial perturbations. 
Taking the camera systems of self-driving cars as an example,  small 
adversarial 
perturbations can cause the system to  make errors in important tasks, such as 
classifying traffic signs or detecting pedestrians.
Hence, in order to use deep learning without safety concerns a proper 
defense strategy is required. 
We propose to use ensemble methods as a defense strategy against adversarial 
perturbations. 
We find that an attack leading one model to misclassify does not imply the same 
for other networks performing the same task.
This makes ensemble methods an attractive defense strategy against adversarial 
attacks. 
We empirically show for the MNIST and the CIFAR-10 data sets that ensemble 
methods not only improve the accuracy of neural networks on test data but also 
increase their robustness against adversarial perturbations.
\end{abstract}

\section{Introduction}

\begin{figure*}[t]
\includegraphics[scale=.14,trim=100px 40px 100px 40px, 
clip]{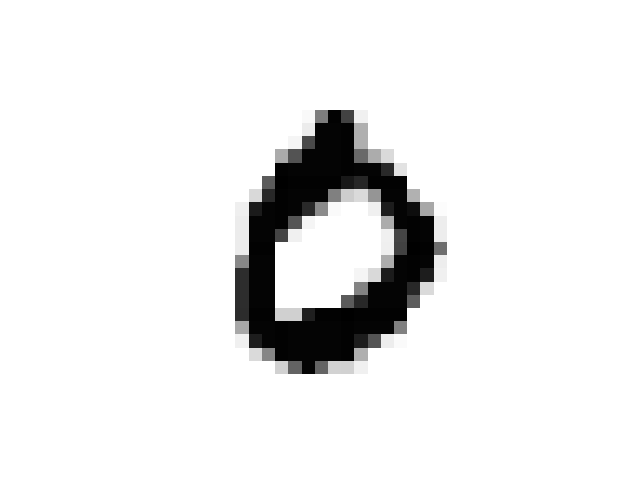}   \hfill 
\includegraphics[scale=.14,trim=100px 40px 100px 40px, 
clip]{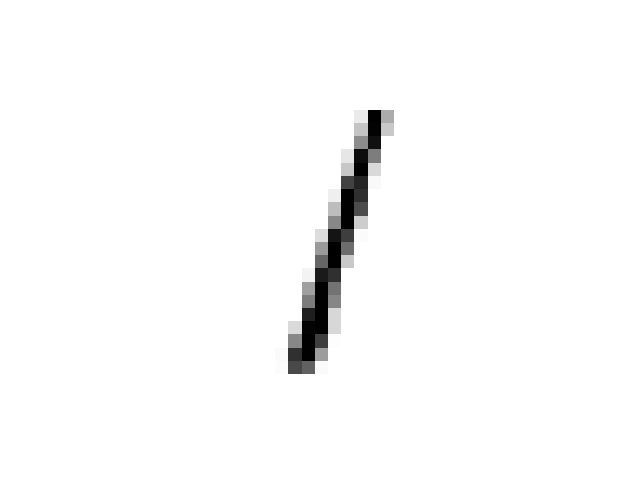}  \hfill  
\includegraphics[scale=.14,trim=100px 40px 100px 40px, 
clip]{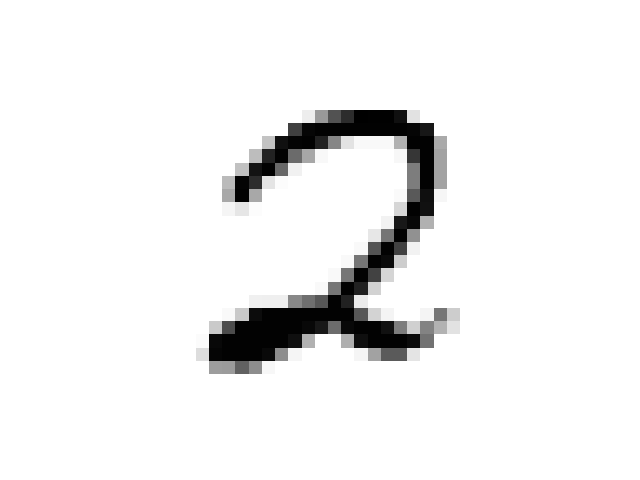}  \hfill 
\includegraphics[scale=.14,trim=100px 40px 100px 40px, 
clip]{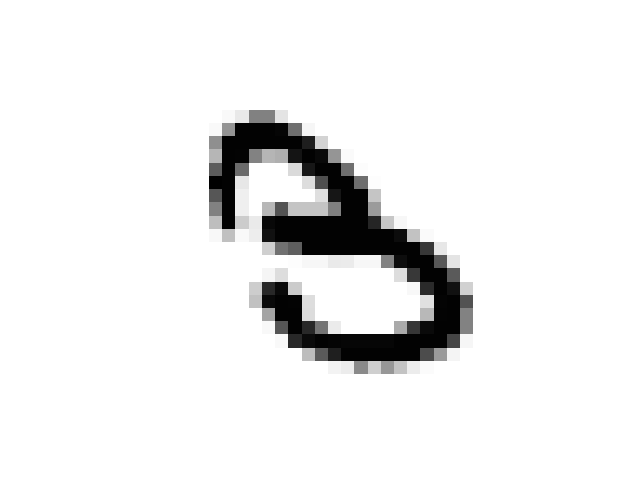}  \hfill  
\includegraphics[scale=.14,trim=100px 40px 100px 40px, 
clip]{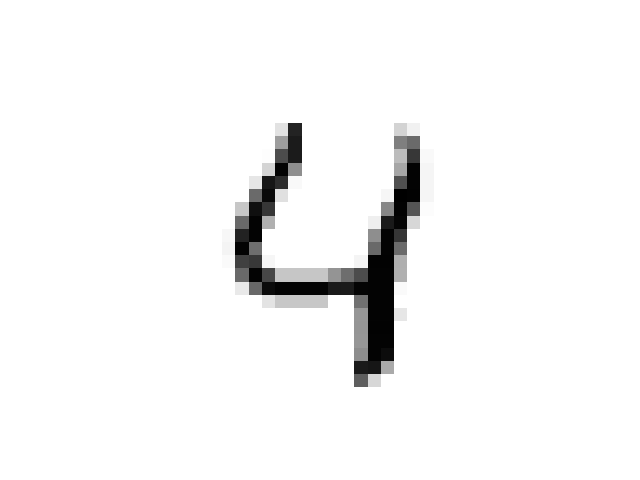} \hfill  
\includegraphics[scale=.14,trim=100px 40px 100px 40px, 
clip]{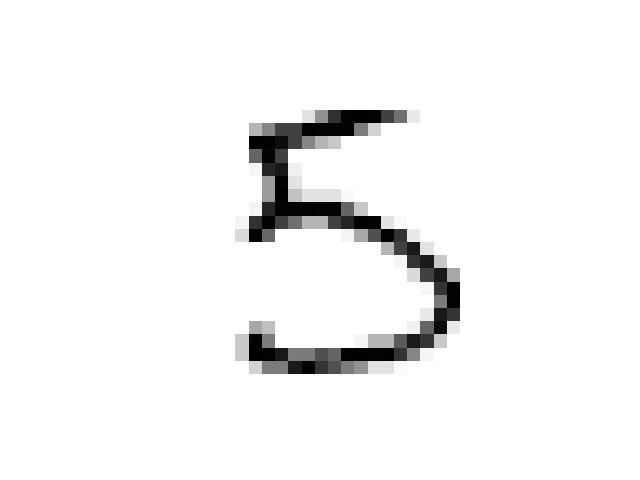}   \hfill 
\includegraphics[scale=.14,trim=100px 40px 100px 40px, 
clip]{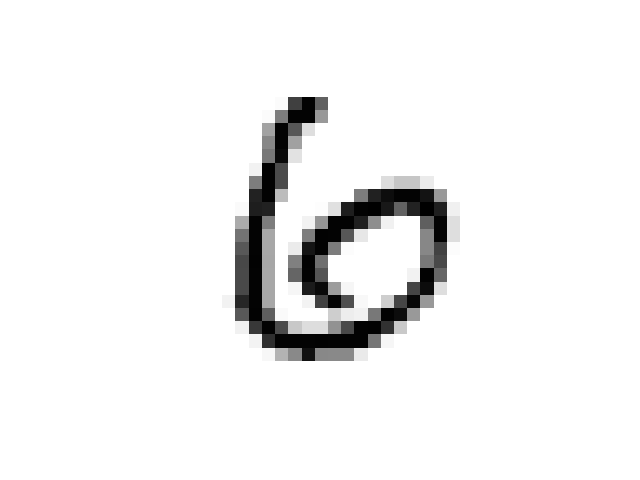}  \hfill  
\includegraphics[scale=.14,trim=100px 40px 100px 40px, 
clip]{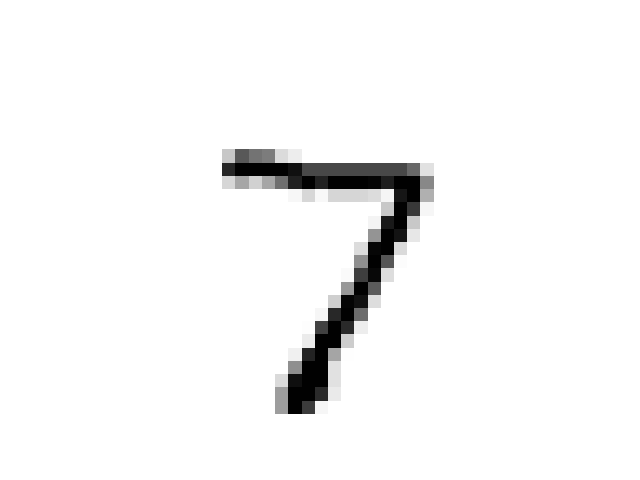}  \hfill  
\includegraphics[scale=.14,trim=100px 40px 100px 40px, 
clip]{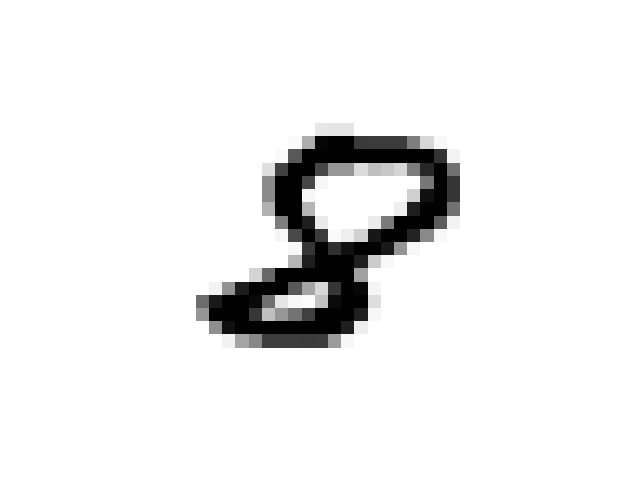}   \hfill 
\includegraphics[scale=.14,trim=100px 40px 100px 40px, 
clip]{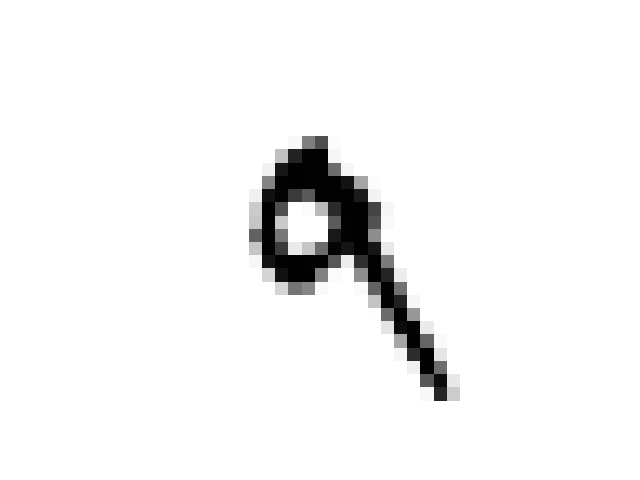}
\\
\includegraphics[scale=.14,trim=100px 40px 100px 40px, 
clip]{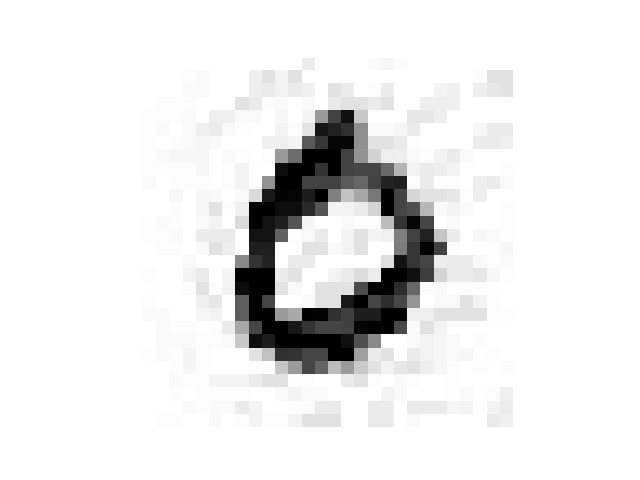} \hfill 
\includegraphics[scale=.14,trim=100px 40px 100px 40px, 
clip]{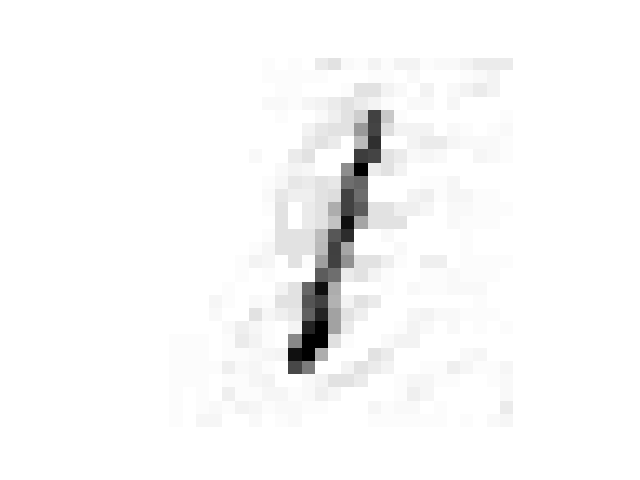}  \hfill 
\includegraphics[scale=.14,trim=100px 40px 100px 40px, 
clip]{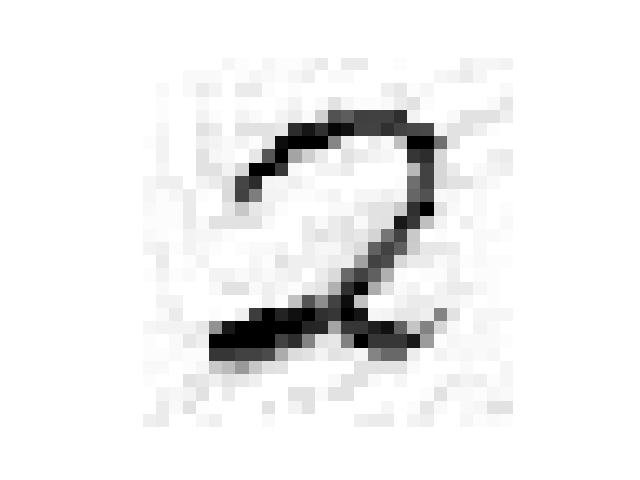}   \hfill  
\includegraphics[scale=.14,trim=100px 40px 100px 40px, 
clip]{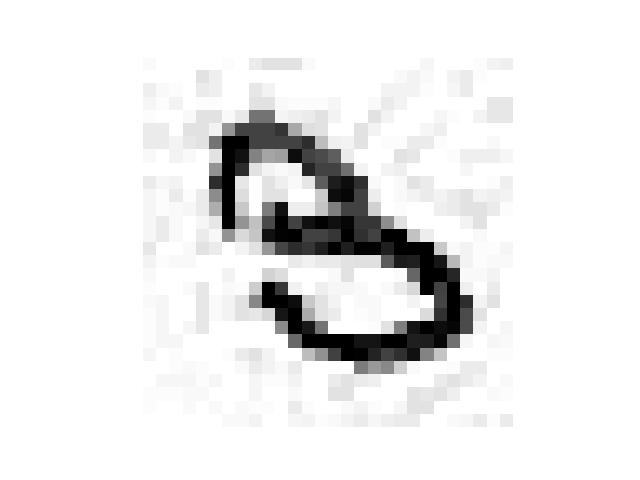}  \hfill  
\includegraphics[scale=.14,trim=100px 40px 100px 40px, 
clip]{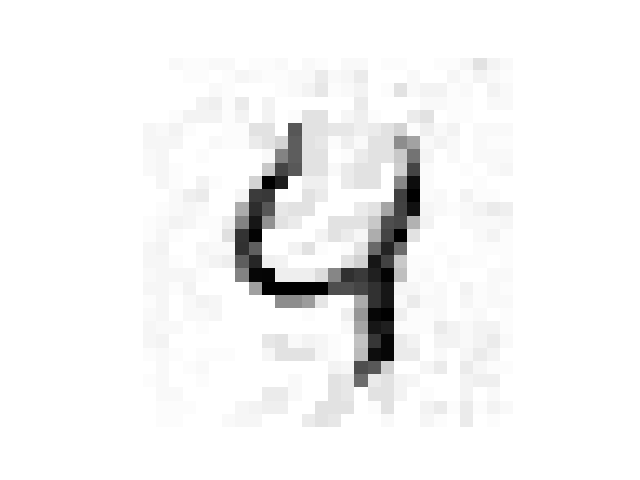} \hfill  
\includegraphics[scale=.14,trim=100px 40px 100px 40px, 
clip]{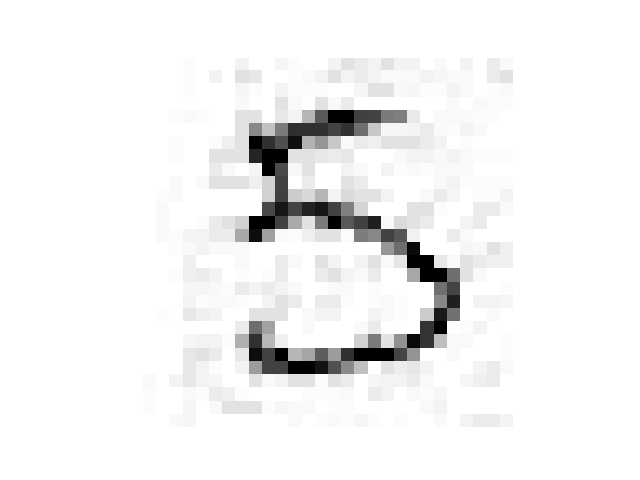}  \hfill  
\includegraphics[scale=.14,trim=100px 40px 100px 40px, 
clip]{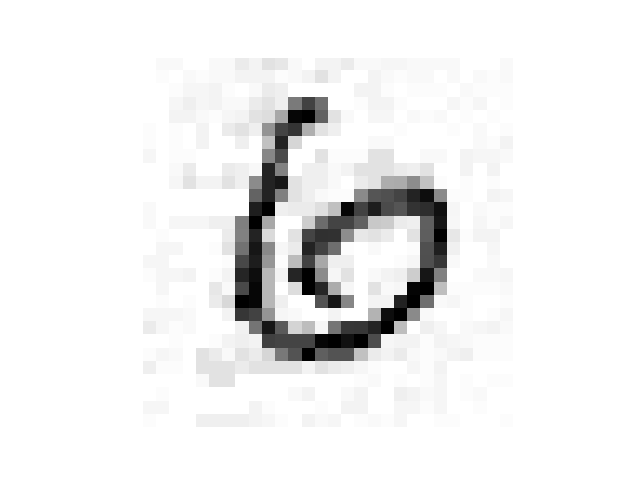}   \hfill  
\includegraphics[scale=.14,trim=100px 40px 100px 40px, 
clip]{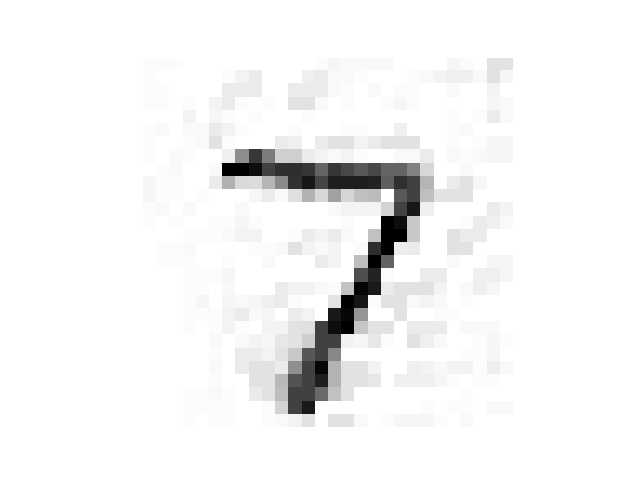}  \hfill 
\includegraphics[scale=.14,trim=100px 40px 100px 40px, 
clip]{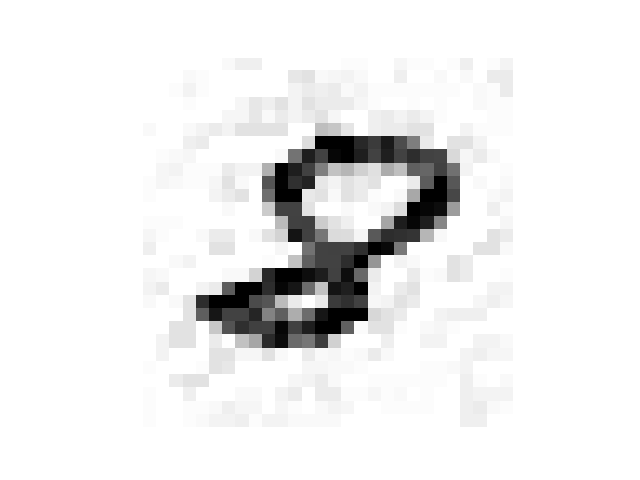}   \hfill  
\includegraphics[scale=.14,trim=100px 40px 100px 40px, 
clip]{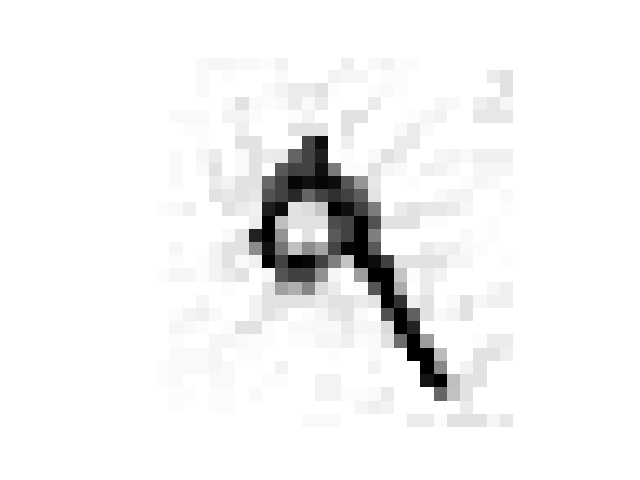}
\\[1ex]
\includegraphics[scale=.14,trim=100px 40px 100px 40px, 
clip]{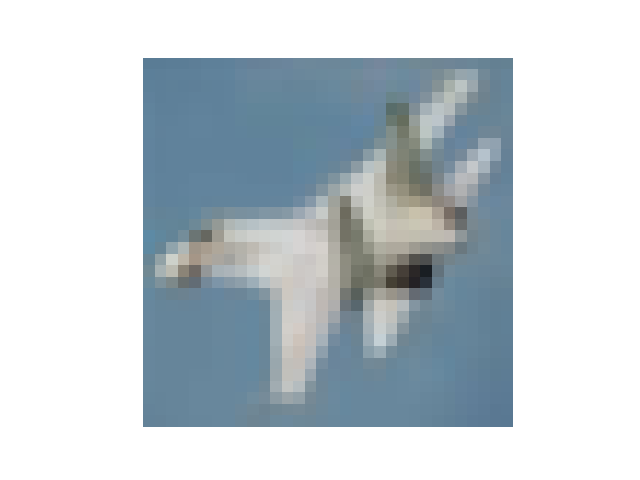} \hfill  
\includegraphics[scale=.14,trim=100px 40px 100px 40px, 
clip]{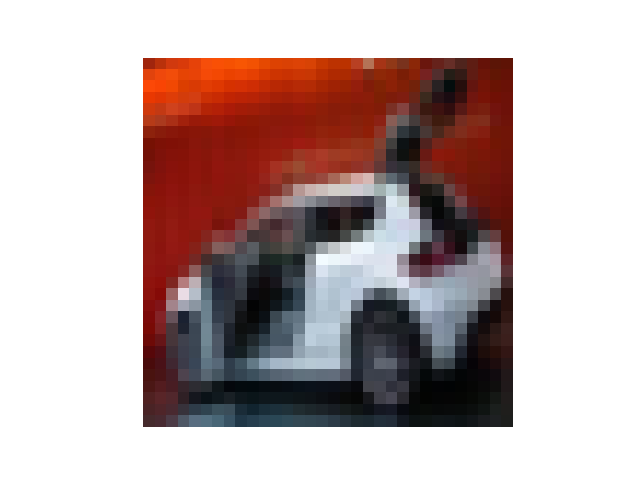}  \hfill  
\includegraphics[scale=.14,trim=100px 40px 100px 40px, 
clip]{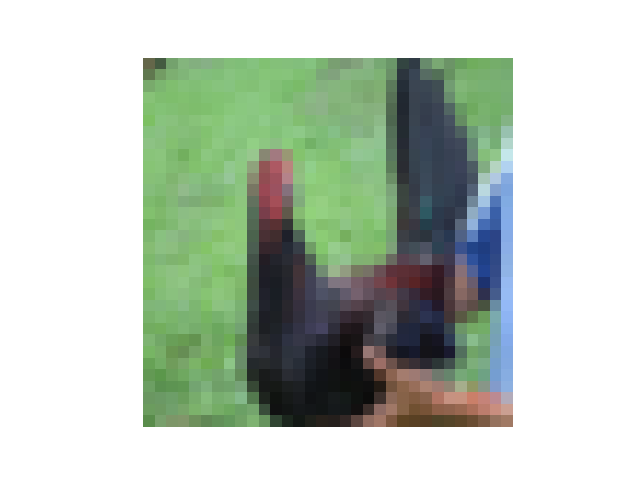} \hfill  
\includegraphics[scale=.14,trim=100px 40px 100px 40px, 
clip]{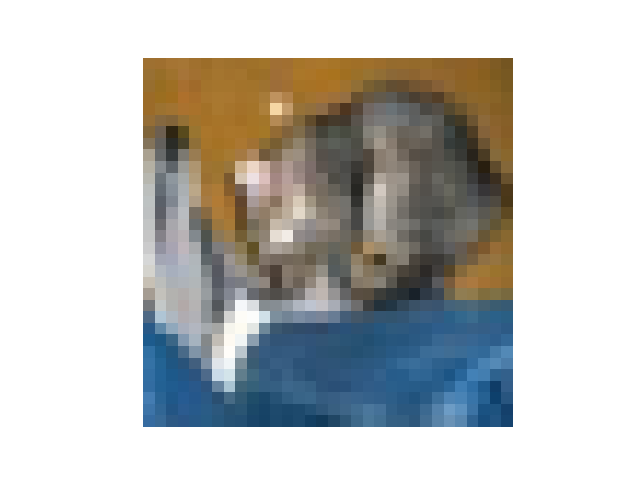}  \hfill  
\includegraphics[scale=.14,trim=100px 40px 100px 40px, 
clip]{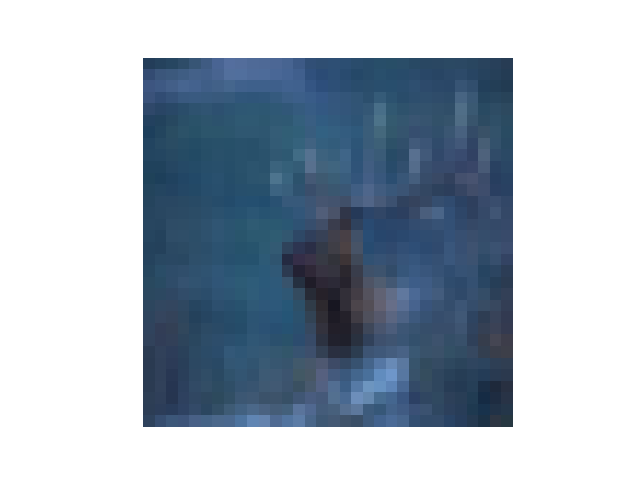} \hfill 
\includegraphics[scale=.14,trim=100px 40px 100px 40px, 
clip]{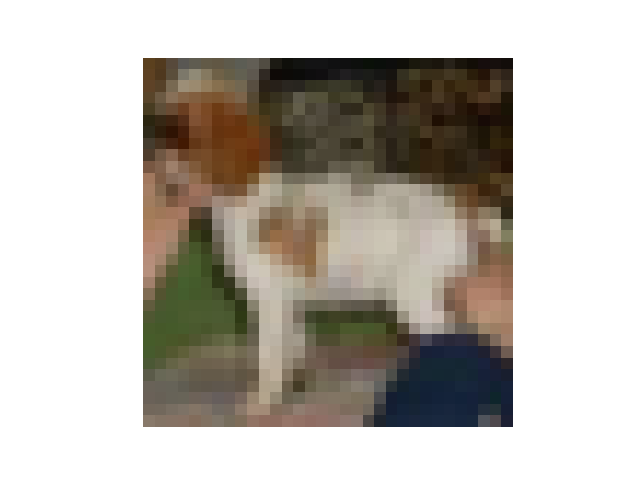} \hfill  
\includegraphics[scale=.14,trim=100px 40px 100px 40px, 
clip]{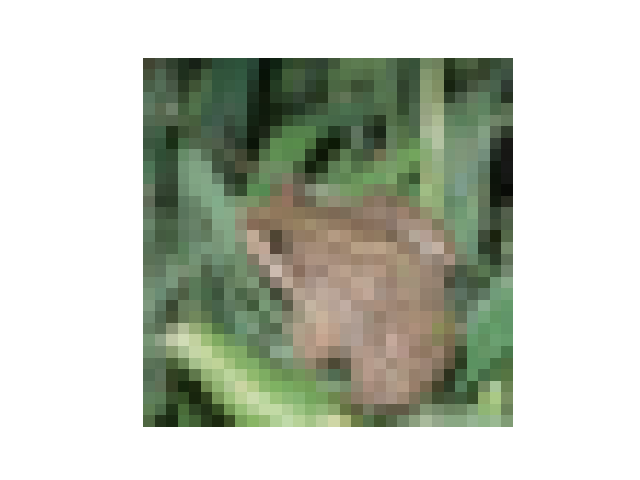}  \hfill  
\includegraphics[scale=.14,trim=100px 40px 100px 40px, 
clip]{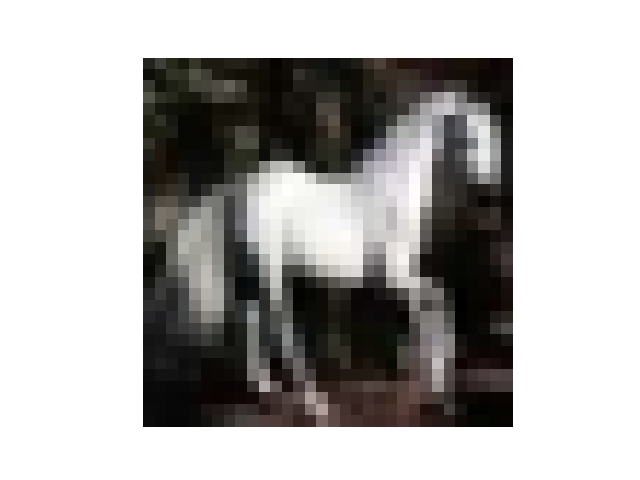} \hfill  
\includegraphics[scale=.14,trim=100px 40px 100px 40px, 
clip]{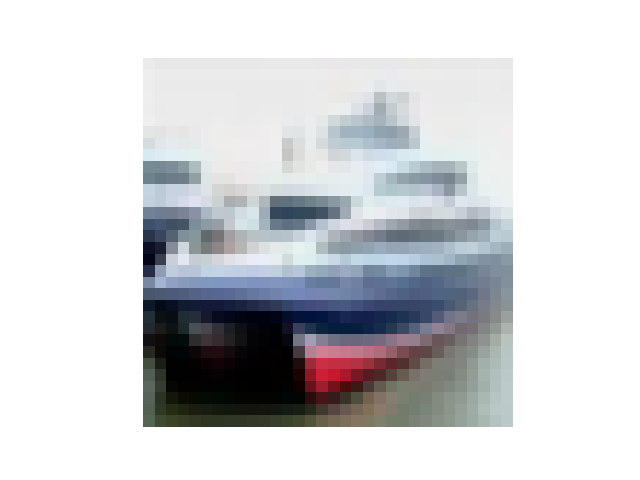}  \hfill 
\includegraphics[scale=.14,trim=100px 40px 100px 40px, 
clip]{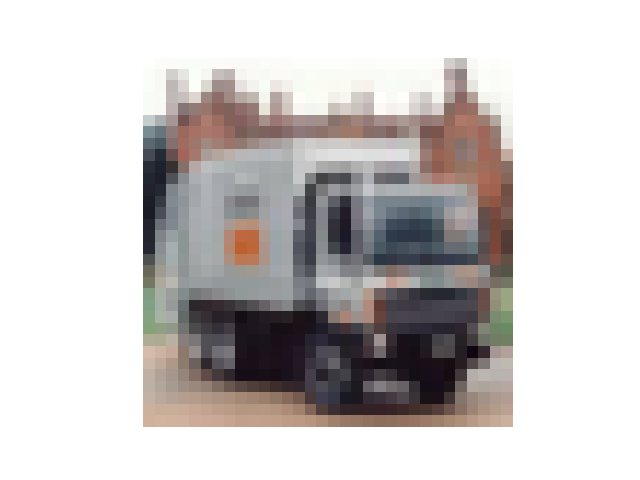}
\\[.9ex]
\includegraphics[scale=.14,trim=100px 40px 100px 40px, 
clip]{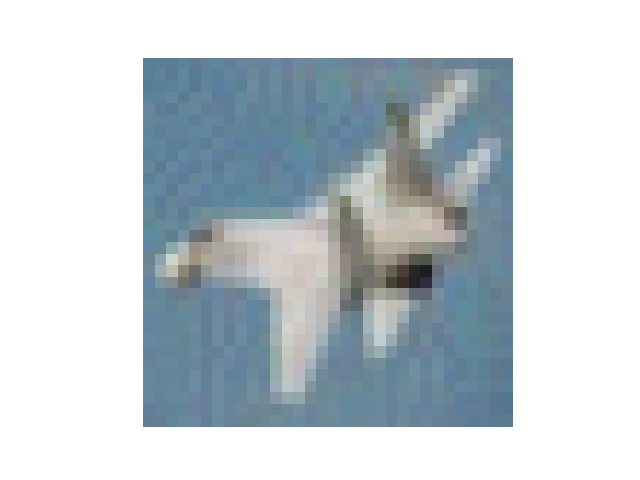}  \hfill  
\includegraphics[scale=.14,trim=100px 40px 100px 40px, 
clip]{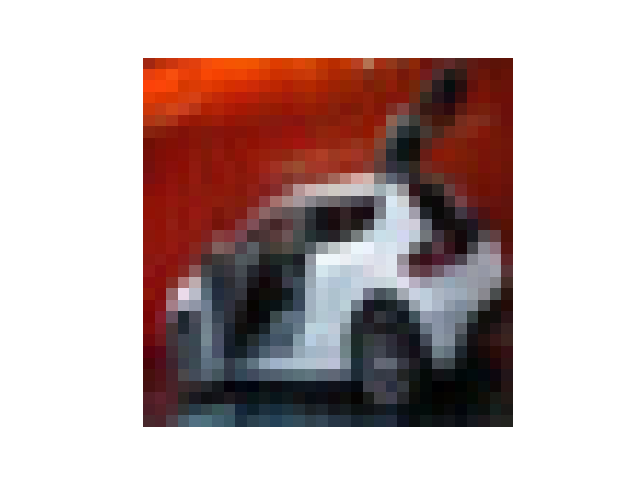}  \hfill  
\includegraphics[scale=.14,trim=100px 40px 100px 40px, 
clip]{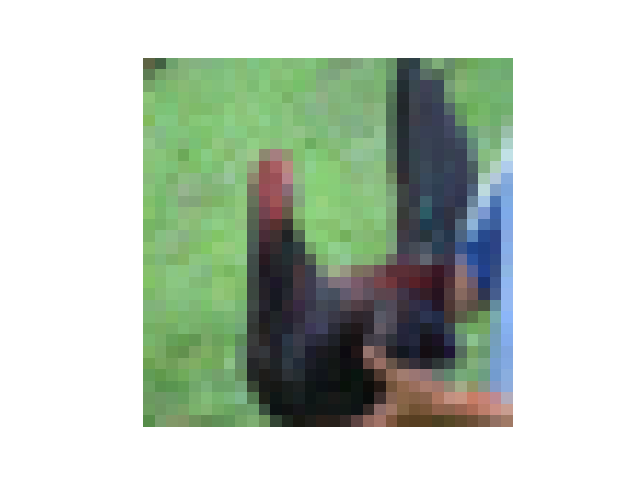}   \hfill
\includegraphics[scale=.14,trim=100px 40px 100px 40px, 
clip]{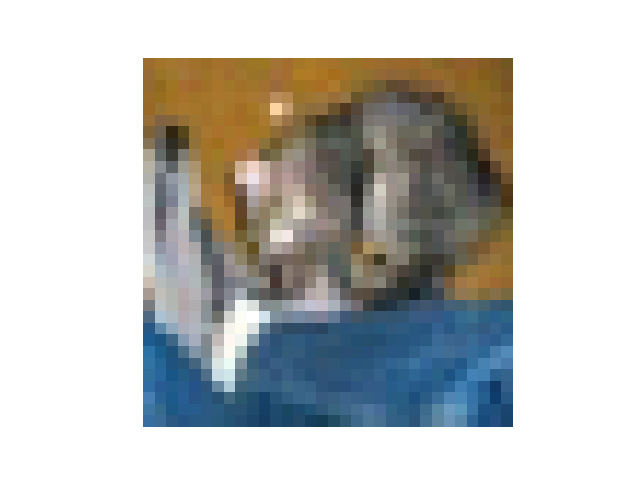} \hfill 
\includegraphics[scale=.14,trim=100px 40px 100px 40px, 
clip]{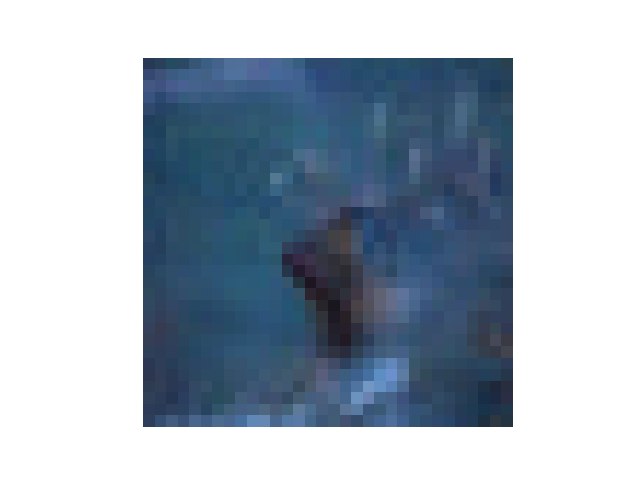} \hfill  
\includegraphics[scale=.14,trim=100px 40px 100px 40px, 
clip]{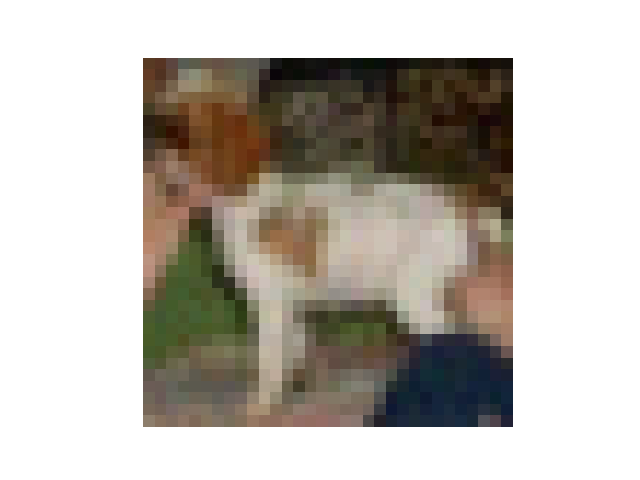} \hfill 
\includegraphics[scale=.14,trim=100px 40px 100px 40px, 
clip]{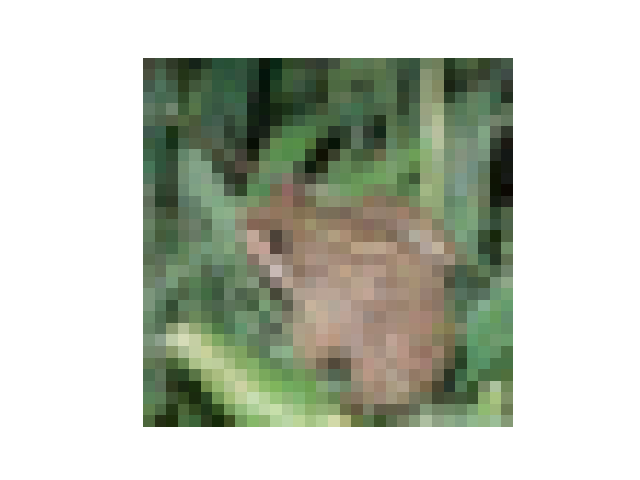} \hfill 
\includegraphics[scale=.14,trim=100px 40px 100px 40px, 
clip]{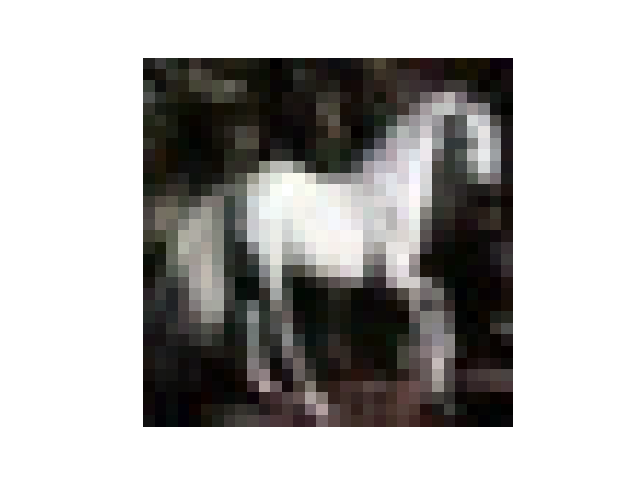} \hfill
\includegraphics[scale=.14,trim=100px 40px 100px 40px, 
clip]{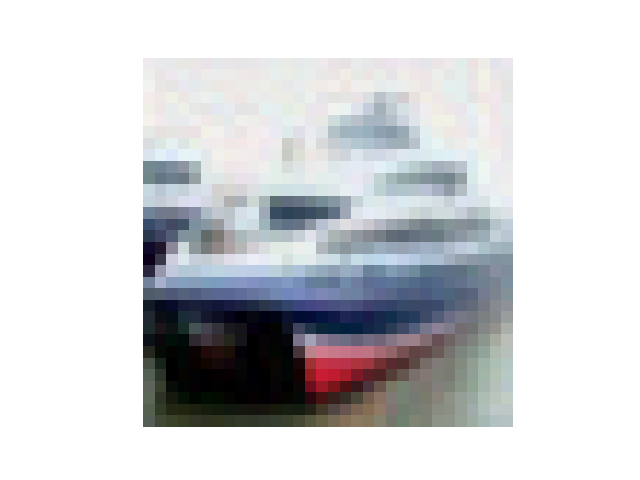}  \hfill  
\includegraphics[scale=.14,trim=100px 40px 100px 40px, 
clip]{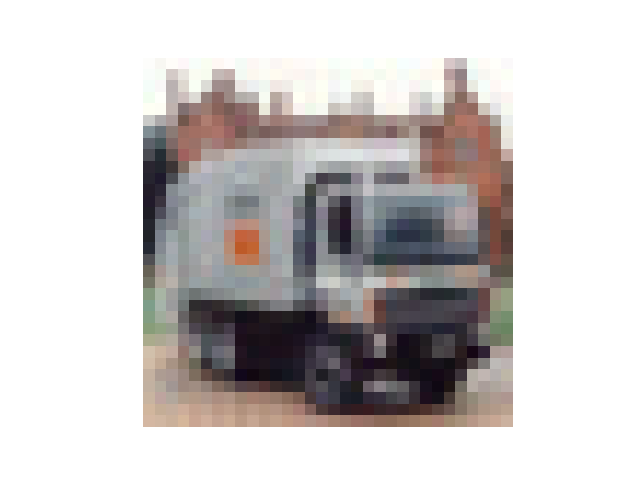} 
%
\caption{%
The first line shows original and correctly classified MNIST test data images. 
In the second line are the corresponding adversarial BIM attacks on a single 
classifier
($\epsilon=0.2$, $\alpha=0.025$, $n=8$) which predicts (from left to 
right): 6, 8, 1, 5, 9, 3, 0, 2, 2, and 4.
Analogously, the third line corresponds to correctly predicted examples of the 
CIFAR-10 test data set. 
In the bottom line are the corresponding adversarial BIM attacks on a single 
classifier
($\epsilon=0.02$, $\alpha=0.0025$, $n=8$) which predicts (from left to 
right):  deer, cat, deer, ship, bird, deer, deer, frog, automobile, and 
automobile.
}
\label{examplePics}
\end{figure*}

In recent years, deep neural networks (DNNs) led to significant improvements in 
many areas ranging from computer vision 
\citep{krizhevsky2012imagenet,lecun2015deep} 
to speech recognition  
\citep{hinton2012deep,dahl2012context}. 
Some applications that can be solved with DNNs are sensitive from the security 
perspective, for example camera systems of self driving cars for detecting 
traffic signs or pedestrians \citep{papernot2016practical,sermanet2011traffic}.
Recently, it has been shown that DNNs can be highly vulnerable to adversaries 
\citep{szegedy2013intriguing,goodfellow2014explaining,papernot2016cleverhans,
papernot2016practical}. 
The adversary produces some kind of noise on the input of the system to mislead 
its output behavior, producing undesirable outcomes or misclassification. 
Adversarial perturbations are carefully chosen in order to be hard, if not 
impossible, to be detected by the human eye (see \FIG~\ref{examplePics}). 
Attacks occur \emph{after} the training of the DNN is completed. 
Furthermore, it has been shown that the exact structure of the DNN does not need 
to 
be known in order to mislead the system as  one can send inputs to the unknown 
system in order to record its outputs to train a new DNN that imitates its 
behavior \citep{papernot2016practical}. 
Hence, in this manuscript it is assumed that the DNN and all its parameters are 
fully known to the adversary.

There are many methods on how to attack neural networks appearing in the 
literature. Some of the most well-known ones are the Fast Gradient Sign Method 
\citep{goodfellow2014explaining}  and its iterative extension 
\citep{kurakin2016adversarial}, DeepFool \citep{moosavi2016deepfool}, 
Jacobian-Based Saliency Map Attack \citep{papernot2016limitations}, C\&W attack \citep{carlini2017towards}, and the  
L-BFGS Attack \citep{szegedy2013intriguing}. 
This shows the need of  building neural networks that are themselves robust 
against any kind of adversarial perturbations.

\def\trimvalue{10px}
\newlength{\trval}
\setlength{\trval}{10px}

Novel methods on defending against adversarial attacks are appearing more and 
more 
frequently in the literature. 
Some of those defense methods  are to train the network with different kinds of 
adversarially perturbated training data  
\citep{goodfellow2014explaining}, 
the use of distillation to reduce the effectiveness of the perturbation 
\citep{papernot2016distillation}
or to apply denoising autoencoders to preprocess the data used by the DNN 
\citep{gu2014towards}. 
It also has been noted that adversarial attacks can be detected 
\citep{metzen2017detecting, feinman2017detecting},
but these detection systems are again vulnerable to adversarial attacks. To our 
knowledge, there is no method that can reliably defend or detect \emph{all} 
kinds 
of adversarial attacks. 

In this manuscript, ensemble methods are used to obtain a classification system 
that is more robust against adversarial perturbations.
The term ensemble method refers to constructing a set of classifiers used
to classify new data points by the weighted or unweighted average of their 
predictions. 
Many ensemble methods have been introduced in the literature such as Bayesian 
averaging, Bagging \citep{breiman1996bagging} and boosting 
\citep{dietterich2000ensemble}. These methods frequently win machine learning 
competitions, for example the Netflix prize \citep{koren2009bellkor}.

Recently, the idea of improving particular defense methods with ensembles 
emerged in the literature. In \citep{abbasi2017robustness} a special kind of 
ensembles of specialist networks was proposed, but it turned out that this 
method could be fooled \citep{he2017adversarial}. Later, it was proposed to 
improve adversarial training by using ensembles of adversarial trained networks 
\citep{tramer2017ensemble}. However, this method reduces the accuracy on 
unperturbed test data. This problem arises in most, if not all, relatively 
successful defense mechanisms against adversarial perturbations (see Table 4).  
With this in mind, most defense methods might not be practical for real life 
applications where state of the art accuracies are required on unperturbed test 
data. 

This is the first paper that considers ensemble methods as sole defense. This 
comes with the advantage that they improve the accuracy on unperturbed test data 
while increasing the robustness against adversarial perturbations considerably. 
To the best of our knowledge, this is the only defense method with these 
properties. The advantages come at the cost of an increase of computational 
complexity and memory requirements.

This paper is organized as follows: 
In \SEC~\ref{attack}, some methods for producing adversarial perturbations are 
briefly introduced.
Section~\ref{Def} describes the defense strategy proposed in this manuscript. 
In \SEC~\ref{Experiments}, the previous methods  are tested on 
the MNIST and CIFAR-10 data sets and are compared to other defense strategies 
appearing in the literature. 
Finally, in \SEC~\ref{Conclusion} the conclusions are presented.

\section{Adversarial Attack}\label{attack}

In this section, two methods for producing adversarial attacks shall  be 
briefly described. In the following, let $\theta$ be the parameters of a model, 
$x$ the input of the model and $y$ the output value associated with the input 
value $x$. Further, let $J(\theta, x, y)$ be the cost function used to train the 
DNN.

\subsection{Fast Gradient Sign Method} 

The fast gradient sign method (\fgsm) by \citet{goodfellow2014explaining} simply 
adds some small perturbations of size 
$\epsilon>0$ to the input $x$,
\begin{equation*}
x_\text{\fgsm} = x + \epsilon \sign [\nabla_{\!x} J(\theta, x, y)] \,,
\end{equation*}
where the gradient $\nabla_{\!x} J(\theta, x, y)$ can be computed using 
backpropagation. 
This relatively cheap and simple adversarial perturbation performs well on many 
DNNs. It is believed that this behavior is due to linear elements such as 
ReLUs or maxout networks in the DNNs  \citep{goodfellow2014explaining}.

\subsection{Basic Iterative Method}

The basic iterative method (BIM) by \citet{kurakin2016adversarial} is an 
iterative 
extension of FGSM. 
The idea is to choose $\epsilon\geq \alpha>0$ and then apply some perturbations
similar to FGSM to the input $x$ and repeat the process $n$ times:
\begin{align*}
x_{0} &= x,\\
x_i &= \clip_{x, \epsilon} \left( x_{i-1} + \alpha 
\sign [\nabla_{\!x_{i-1}} J(\theta, x_{i-1}, y)]\right),\\
x_\text{BIM} &= x_n.
\end{align*}
Here, $\clip_{x, \epsilon}(\cdot)$ refers to clipping the values of the 
adversarial sample so that they remain within an $\epsilon$-neighborhood of $x$.

\section{Ensemble Methods}\label{Def}

Ensemble methods are widely used to improve classifiers in supervised 
learning \citep{dietterich2000ensemble}. The idea is to construct a set of 
classifiers that is used to classify a new data point  by the weighted or 
unweighted average of their predictions. 
In order for an ensemble to outperform  a single classifier it must be both 
accurate and diverse \citep{hansen1990neural}.
A classifier is said to be accurate if it is better than random guessing, and 
a set of classifiers is said to be diverse if different classifiers make 
different errors on new data points.

As expected, when performing adversarial perturbations on new data points 
different classifiers perform quite 
differently on these points. 
Hence, we conclude that diversity on adversarial perturbations is given. 
Furthermore, for adversarial perturbations with small $\epsilon>0$, the vast  
majority of classifiers was accurate. 
In other words, for any small $\epsilon>0$, we could not find an adversarial 
attack that would turn the majority of classifiers into non-accurate 
classifiers.

In section \ref{Experiments}, the following ensemble methods 
are used. Note that random initialization of the model parameters is used in all 
methods.
\begin{enumerate}[(i)]
\item The first method is  to train multiple classifiers with the same network 
architecture but with random initial weights. 
This results in quite diverse classifiers with different final weights  
\citep{kolen1991back}. 

\item The second method is to train multiple classifiers with different but 
similar 
network architectures to ensure obtaining a set of even more diverse 
classifiers. 
That is, extra filters are used in one classifier or an extra 
convolution layer is added to another classifier.

\item Third, \emph{Bagging} \citep{breiman1996bagging} is used on the training 
data. 
The term Bagging is derived from bootstrap aggregation and it consists of 
drawing $m$ samples with replacement from the training data
set of $m$ data points. 
Each of these new data sets is called a bootstrap replicate. 
At average each of them contains $63.2\%$ of the training data, where many data 
points are repeated in the bootstrap replicates. 
A different bootstrap replicate is used as training data for each classifier in 
the ensemble.  

\item The last method is to add some small Gaussian noise to the training data 
so 
that all classifiers are trained on similar but different training sets. 
Note that adding Gaussian noise to the training data also makes each classifier 
somewhat more robust against adversarial perturbations.
\end{enumerate}

\begin{table}[t]
\caption{MNIST Network Architecture}
\begin{center}
    \begin{tabular}{llp{5cm}}
\toprule
   Layer Type  & Parameters   \\ \midrule
   Relu Convolutional & 32 filters (3$\times$3)  \\ 
   Relu Convolutional & 32 filters (3$\times$3)  \\ 
   Max Pooling & 2$\times$2  \\ 
   Relu Fully Connected & 128 units  \\ 
   Dropout & 0.5 \\ 
   Relu Fully Connected & 10 units \\ 
 Softmax & 10 units \\ 
\bottomrule
    \end{tabular}
\end{center}
\label{MNIST_struc}
\end{table}

Once an ensemble of classifiers is trained, it predicts by letting each 
classifier vote for a label. 
More specifically, the predicted value is chosen to be the label that maximizes 
the average of the output probabilities from the classifiers in the ensemble.

In order to attack a network with the methods from \SEC~\ref{attack} the 
gradient $\nabla_x J(\theta, x, y)$ must be computed. To evaluate 
the robustness of the proposed ensemble methods, two 
different gradients are used. They are referred to as Grad. 1 and Grad. 2:

\begin{description}
\item[Grad.~1]
Use $\nabla_x J(\theta_i, x, y)$ of the $i$-{th} classifier. This is clearly not 
the correct gradient for an ensemble. But the question is whether an attack with 
this gradient can already mislead all classifiers in the ensemble in a similar 
manner.

\item[Grad.~2]
Compute the average of the gradients $\frac{1}{n}\sum_i\nabla_xJ(\theta_i, x, 
y)$ from all classifiers in the ensemble. 

\end{description}

A comparison of the effects of these two gradients for attacking ensembles can 
be found in \SEC~\ref{Experiments}.

\section{Experiments}\label{Experiments}

\begin{table*}[t]
\centering
\caption{Experimental results on the MNIST and the CIFAR-10 data 
sets} 
\begin{tabular}{lllllll}
\multicolumn{7}{c}{MNIST Accuracy}\\[1ex]
  \toprule
  \multicolumn{2}{c}{Test Data}   & \multicolumn{2}{c}{No Attack}  & 
\multicolumn{2}{c}{Grad. 1}&Grad. 2\\
  \midrule
  Type&Method  &Single& Ensemble  &Single& Ensemble  & Ensemble\\   \midrule
\multirow{3}{*}{FGSM} &Rand. Ini.  & 0.9912 & 0.9942  & 0.3791 & 0.6100  & 
0.4517  \\  
 &Mix. Mod.&0.9918&0.9942&0.3522&0.5681&0.4609 \\  
&Bagging  & 0.9900 & 0.9927 &  0.4045 & 0.6738  & 0.5716 \\ 
&Gauss Noise & 0.9898 & 0.9920  & 0.5587  & 0.7816  & 0.7043 \\ \midrule
\multirow{3}{*}{BIM} &Rand. Ini.  &0.9912&0.9942&0.0906&0.6518 &0.8875\\  
&Mix. Mod.  & 0.9918 & 0.9942 & 0.0582 & 0.6656 & 0.9076 \\ 
&Bagging & 0.9900 & 0.9927 &0.1110 & 0.7068  & 0.9233\\ 
 &Gauss Noise   &0.9898&  0.9920 &0.5429& 0.9152 & 0.9768\\ \bottomrule \\[2em]
%
%
\multicolumn{7}{c}{CIFAR-10 Accuracy}\\[1ex]
  \toprule
  \multicolumn{2}{c}{Test Data}   & \multicolumn{2}{c}{ No Attack}  & 
\multicolumn{2}{c}{Grad. 1}&Grad. 2\\
  \midrule
  Type&Method  &Single& Ensemble  &Single& Ensemble  & Ensemble\\   \midrule
\multirow{3}{*}{FGSM} &Rand. Ini.  & 0.7984& 0.8448  & 0.1778 &0.4538  & 0.3302 
\\  
&Mix. Mod.&  0.7898 & 0.8400   &  0.1643  & 0.4339  & 0.3140\\  
&Bagging & 0.7815 & 0.8415 & 0.1822 & 0.4788  &  0.3571 \\
 &Gauss Noise   &0.7160& 0.7687   &0.2966& 0.6097 & 0.4707 \\ \midrule
\multirow{3}{*}{BIM} &Rand. Ini.  &0.7984&0.8448&0.1192& 0.5232&0.6826\\  
&Mix. Mod.  & 0.7898 & 0.8400 & 0.1139 & 0.5259 & 0.6768 \\ 
&Bagging & 0.7815 & 0.8415 &0.1280 & 0.5615  &0.7166\\
 &Gauss Noise   &0.7160& 0.7687   &0.3076&  0.6735&0.7277 \\  \bottomrule 
\\[1em]
\end{tabular}
\label{MNIST_res}
\end{table*}

In this section the ensemble methods from \SEC~\ref{Def} are empirically 
evaluated on the MNIST \citep{lecun1998gradient} and the CIFAR-10  
\citep{krizhevsky2009learning} data sets which are scaled to the unit interval. 
All experiments have been performed on ensembles of 10 classifiers. Note that 
this choice has been done for comparability. That is, in some cases the best 
performance was already reached with ensembles of less classifiers while in 
others more classifiers might improve the results. 

\begin{figure*}[t]
\begin{center}
\includegraphics[width=.4\textwidth]{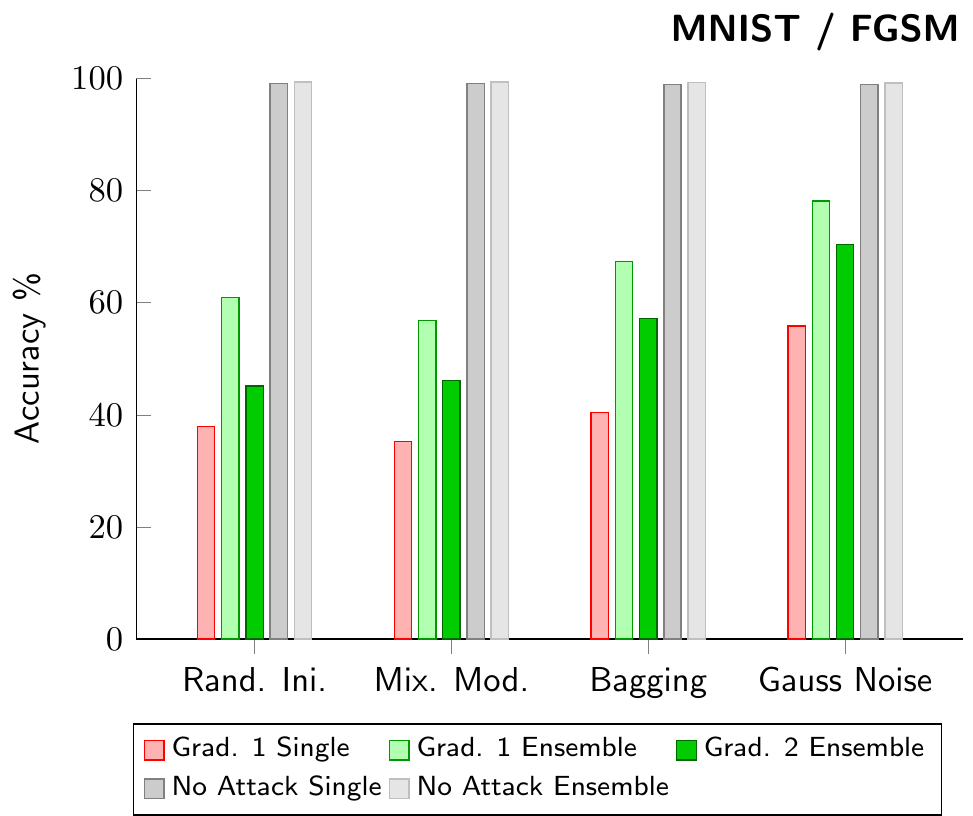} \hspace{3em}
\includegraphics[width=.4\textwidth]{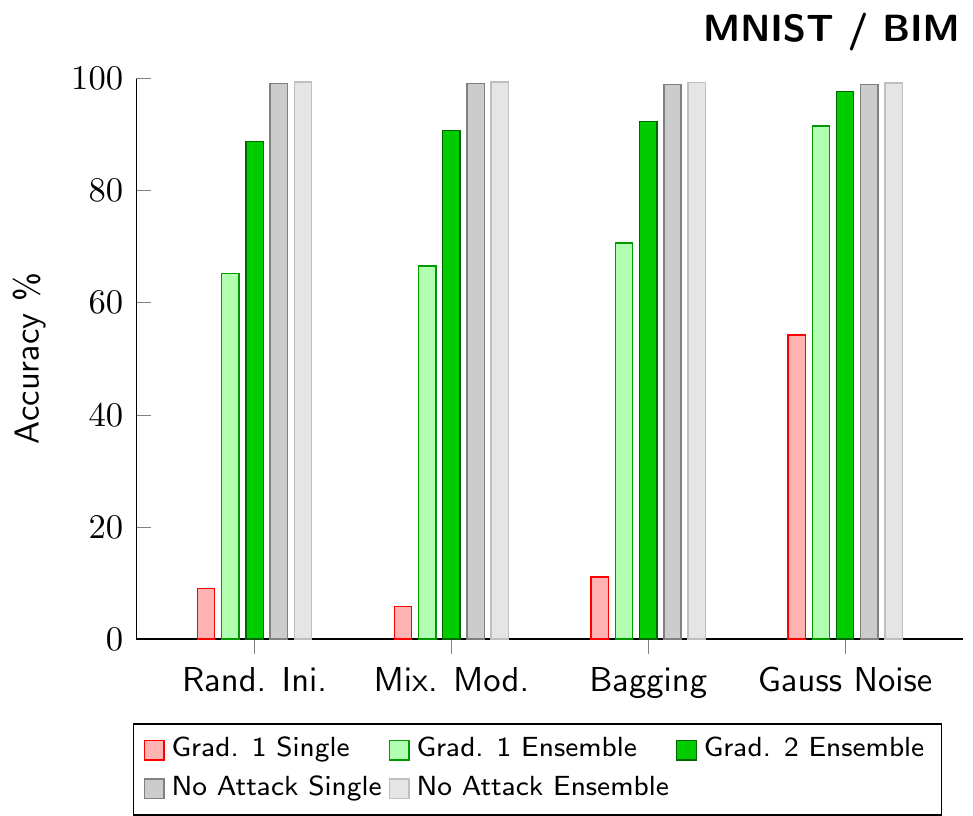} 
\\
\includegraphics[width=.4\textwidth]{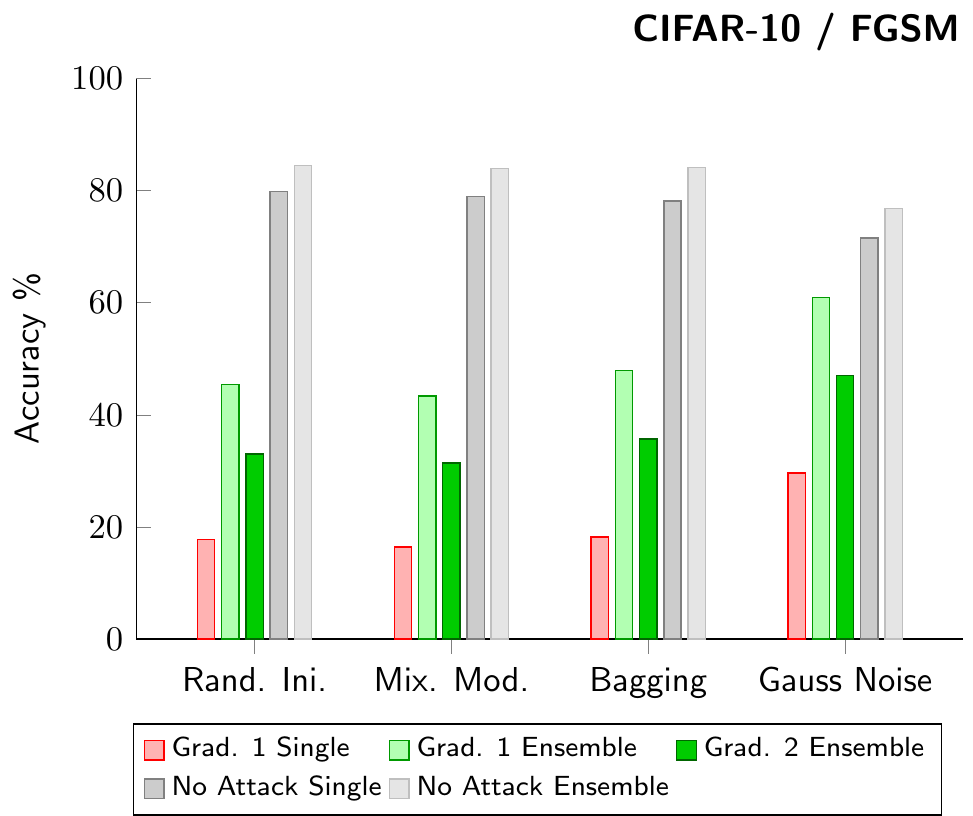} \hspace{3em}
\includegraphics[width=.4\textwidth]{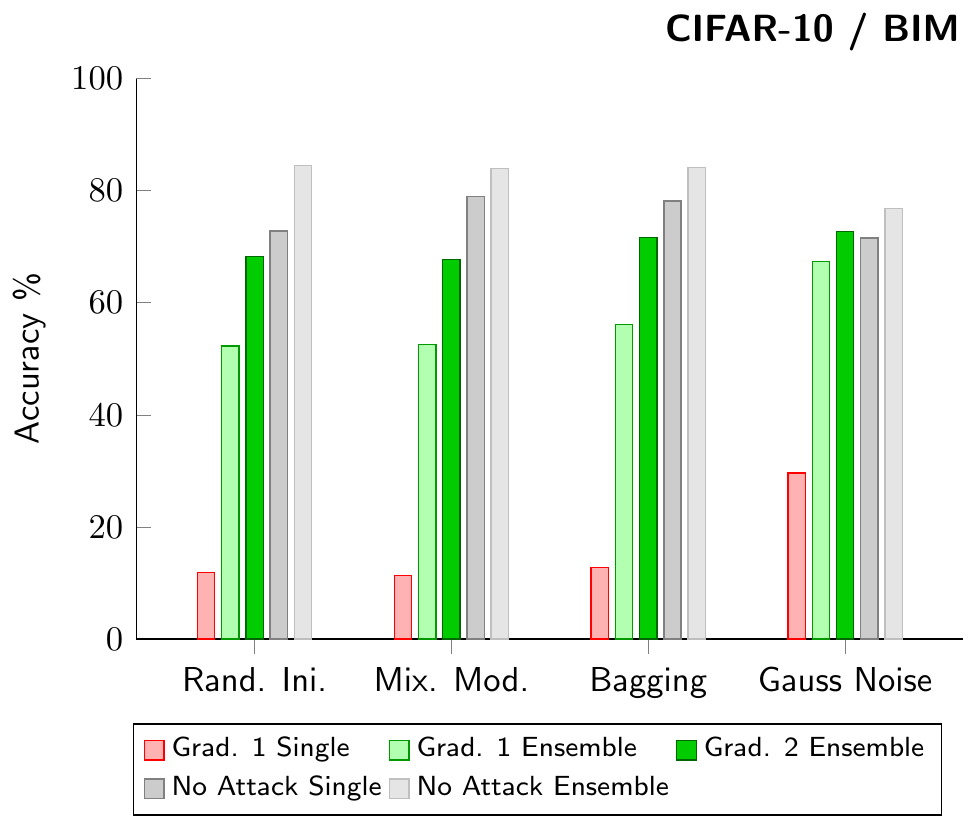} 
\end{center}
\caption{%
Visual comparisons of the accuracies presented in Table~\ref{MNIST_res}.
Compared are the MNIST (top row) and CIFAR-10 (bottom row) data sets on the FGSM 
(left column) and the BIM (right column) attacks.
Grad.~1 Single refers to attacks based on  Grad.~1 on single classifiers, 
Grad.~1 Ensemble refers  to attacks based on  Grad.~1 on ensembles, Grad.~2 
Ensemble refers to attacks based on  Grad.~2 on ensemble classifiers, No Attack 
Single refers to single classifier on unperturbed data, and finally No Attack 
Ensemble refers to ensemble classifiers on  unperturbed data.
}
\label{fig:barcharts}
\end{figure*}

A summary of the experimental results can be found in Table \ref{MNIST_res} and 
the corresponding visualization in Figure \ref{fig:barcharts}. A comparison of 
ensembles with other defense methods and a combination of those with ensembles 
can be found in Table \ref{compare}.
In the following all FGSM perturbations are done with $\epsilon=0.3$ on MNIST 
and with 
$\epsilon=0.03$ on CIFAR-10. 
Furthermore, all BIM perturbations are done with 
$\epsilon=0.2$, $\alpha=0.025$ and $n=8$ iterations on MNIST and with  
$\epsilon=0.02$, $\alpha=0.0025$ and 
$n=8$ on CIFAR-10. The abbreviations in Table \ref{MNIST_res} and in Figure 
\ref{fig:barcharts} shall be interpreted in the following way: Rand.~Ini. refers 
to random 
initialization of the weights of the neural network,  Mix.~Mod. means that 
the network architecture was slightly different for each classifier in an 
ensemble, Bagging refers to classifiers  trained on bootstrap replicates of 
the training data, and Gauss noise implies that small Gaussian noise has 
been added to the training data. Each ensemble is attacked with FGSM and BIM 
based on the gradients from Grad. 1 and Grad. 2. In Table~\ref{MNIST_res}, the 
term Single refers to evaluating a single classifier.

\subsection{MNIST}

The MNIST data set consists of 60,000 training and 10,000 test data samples of 
black and white encoded handwritten digits. 
The objective is to classify these digits in the range from 0 to 9. 
A selection of images from the data set and some adversarial perturbations can 
be found in the top two rows of \FIG~\ref{examplePics}. 
In the experiments, the network architecture in Table \ref{MNIST_struc} is 
used and it is trained with 10 epochs. All results from the experiments are 
summarized in Table \ref{MNIST_res}.

On unperturbed test data the classification accuracy is roughly 99\%.
The difference between single classifiers and ensembles is below one percent 
throughout. 
The ensembles slightly outperform the single classifiers in all cases.

This picture changes dramatically if the networks are attacked by one of the 
methods described in \SEC~\ref{attack}.
Using the FGSM attack with gradients from Grad.~1 on a single classifier, the 
classification rate drops down to a range of roughly 35\%--56\%.
The ensembles perform significantly better by producing an accuracy of 
57\%--78\%.
Evaluating the same with gradients from Grad.~2 it turns out that ensemble 
methods still obtain an accuracy of 45\%--70\%.
The higher accuracy of Grad.~1 is expected since in contrast to Grad.~2 it 
computes the gradients with respect to just one classifier.
Nevertheless, the ensembles outperform single classifiers in each case by 
approximately 7\%-22\%. 

The decrease of the accuracy is even more extreme for single classifiers  if the 
BIM method is 
used.
Here, the accuracy can be as low as around 6\% and only the classifiers trained 
with Gaussian noise significantly exceed the 10\%.
The accuracy of the ensemble methods against attacks using Grad.~1 is 
considerably higher with 65\%--92\%.
Furthermore, ensembles are even more robust against  BIM attacks based on 
Grad.~2 with a correct classification rate 
of 89\%--98\%. It is surprising that BIM attacks using Grad.~1 are more 
successful than those using Grad.~2, because Grad.~1 only attacks a single 
classifier in the ensemble. 
Concluding, the ensemble methods outperform single classifiers significantly by 
37\%-85\% on BIM attacks.

Focusing on the different defense strategies, we observe that using random 
initialization of the network weights as well as using several networks of 
similar architectures for an ensemble generally improves the robustness against 
adversarial attacks considerably  in comparison with single classifiers. 
Bagging outperforms both of the previous methods on adversarial 
perturbations, but performs slightly worse on unperturbed test data. 
Using ensembles with small Gaussian noise on the training data results in the 
best defense mechanism against adversarial attacks.
This may be due to the fact that using additive noise on the training data 
already makes 
every single classifier in the ensemble more robust against adversarial 
perturbations. 
On the down-side, adding Gaussian noise to the training data performs worst 
from all considered ensemble methods on test data. 
However, such an ensemble still performs better than all single classifiers on 
MNIST.

\subsection{CIFAR-10}

The CIFAR-10 data set consists of 50,000 training and 10,000 test data samples 
of 
three-color component encoded images of ten mutually exclusive classes: 
airplane, automobile, bird, cat, deer, dog, frog, horse, ship, and truck. 
A selection of images from the data set and some adversarial perturbations can 
be found in the two bottom rows of \FIG~\ref{examplePics}. 
In all experiments the network architecture described in 
Table \ref{abc} is used and the networks are trained with 25 
epochs. 

In general, the observations on the MNIST data set are confirmed by the 
experiments on CIFAR-10.
Since the latter data set is more demanding to classify, the overall 
classification rate 
is already lower in the attack-free case, where single classifiers reach an 
accuracy of roughly 72\%--80\%, while ensembles show a higher accuracy of 
77\%--84\%. Note that there are network architectures in the literature that 
outperform our classifiers considerably on test data 
\citep{graham2014fractional}.

The FGSM attacks on single classifiers using method Grad.~1 show a drop-down of 
the accuracy 
to 16\%-30\%. In contrast, ensembles are significantly better reaching 
accuracies of 43\%-61\% when attacked using Grad.~1 and 31\%-47\% when attacked 
with Grad.~2.

When using BIM attacks accuracies for single classifiers lie between 11\% and 
31\%. Again, the ensemble methods outperform the single classifiers reaching 
accuracies of 52\%-67\% when attacked using Grad.~1 and 68\%-73\% when attacked 
with Grad.~2. 

The same observations as on the MNIST data set can be made on the CIFAR-10 data 
set. All ensemble methods outperform single classifiers when comparing their 
robustness against adversarial perturbations.
FGSM attacks on an ensemble using Grad.~2 outperform those using Grad.~1, as 
expected. Similar to the MNIST experiments, when using BIM attacks, ensembles 
are surprisingly more robust against gradient attacks from Grad.~2 than against 
gradient attacks from Grad.~1. The reason for this might be that the gradient 
portion from different classifiers using Grad.~2 in the ensemble try to reach a 
different  local maximum and block each other in the following iterations.

As  already observed on the MNIST data set, Bagging performs better than 
random initialization and than using similar but different network 
architectures. 
Again, adding small Gaussian noise on the training data performs best on 
adversarial perturbations but relatively poor on real test data on CIFAR-10.

\begin{table}[t]
\begin{center}
\caption{CIFAR-10 Network Architecture\label{abc}}
    \begin{tabular}{llp{5cm}}
\toprule
   Layer Type  &  Parameters     \\ \midrule
   Relu Convolutional & 32 filters (3$\times$3)\\ 
    Relu Convolutional & 32 filters (3$\times$3)  \\ 
   Max Pooling & 2$\times$2 \\ 
 Dropout & 0.2 \\ 
    Relu Convolutional & 64 filters (3$\times$3) \\ 
    Relu Convolutional & 64 filters (3$\times$3) \\ 
   Max Pooling & 2$\times$2 \\ 
Dropout & 0.3 \\ 
    Relu Convolutional & 128 filters (3$\times$3)  \\ 
    Relu Convolutional & 128 filters (3$\times$3) \\ 
   Max Pooling & 2$\times$2  \\ 
Dropout & 0.4 \\ 
 Relu Fully Connected & 512 units\\ 
      Dropout & 0.5\\ 
   Relu Fully Connected & 10 units  \\ 
 Softmax & 10 units \\ 
 \bottomrule
    \end{tabular}
\end{center}
\end{table}

\subsection{Comparison with other Methods}

In this section, we compare the previous results with two of the most popular 
defense methods: adversarial training
\citep{goodfellow2014explaining}
and defensive distillation \citep{papernot2016distillation}. 
Furthermore, we show the positive effects of combining those methods with 
ensembles. 
For simplicity, we only consider the gradient Grad.~2 whenever an ensemble is 
attacked. 
The results are summarized in Table~\ref{compare}. Here, the content shall be 
interpreted in the following way: Bagging refers to 
ensembles trained with bagging, Adv.~Train.\ to adversarial training, 
Def.~Dist.\ to defensive distillation, the operator $+$ to combinations of 
the previous methods, bold text to the best performance of 
the first three methods, and the asterisk to the best method including 
combinations of defensive strategies. 

Adversarial training (AT) is a method that uses FGSM as regularizer of the 
original cost function:
\begin{equation*}
J_{AT}(\theta, x, y) = \rho J(\theta, x, y)  + (1-\rho) J(\theta, x + 
\epsilon 
\sign(\nabla_{\!x} J(\theta, x, y)) , y),
\end{equation*}
where $\rho \in [0,1]$. This method iteratively increases the robustness against 
adversarial perturbations. In our experiments, we use $\rho=\frac{1}{2}$ as 
proposed in \cite{goodfellow2014explaining}.

In defensive distillation a teacher model $F$ is trained on a training data set 
$X$. Then smoothed labels at temperature $T$ are computed by 
\begin{align*}
F^T(X)= \left[ \frac{\exp(F_i(X)/T)}{\sum_{i=1}^N\exp(F_i(X)/T)}\right]_{i\in 
\{1, ...,N\}},
\end{align*}
where $F_i(X)$ refers to the probability of the $i$-th out of $N$ possible 
classes. A distilled network is a network that is  trained on the training data 
$X$ using the smoothed labels  $F^T(X)$. In the following, we use $T=10$ based 
on the experimental results in \cite{papernot2016distillation}.

We found that single networks trained with adversarial training or defensive 
distillation have a lower accuracy than ensembles trained with bagging (see the 
top three rows in Table \ref{compare}). This is not only the case on the 
considered attacked data but also on unperturbed test data. 
Combining ensembles with adversarial training can improve the robustness against 
adversarial perturbations further, while a combination with defensive 
distillation does not reveal the same tendency (see the two bottom rows in Table 
\ref{compare}).
We emphasize that already the standard ensemble method does not only outperform 
both adversarial training and defensive distillation throughout but also has the 
overall highest accuracy on unperturbed test data.

\begin{table*}[t]
\centering
\caption{Accuracies of different defense mechanisms} 
\begin{tabular}{l l l l l l l}
  \toprule
    & \multicolumn{3}{c}{MNIST}  & 
\multicolumn{3}{c}{CIFAR-10}\\
  \midrule
Methods & No Attack & FGSM & BIM &No Attack & FGSM & BIM  \\ 
  \midrule
 Bagging & \textbf{0.9927}$^*$  & \textbf{0.5716} & \textbf{0.9233} & 
\textbf{0.8415}$^*$ & \textbf{0.3571} & \textbf{0.7166}$^*$  \\
 Adv. Train. & 0.9902 & 0.3586 & 0.5420 & 0.7712 & 0.1778 & 0.3107 \\
 Def. Dist. & 0.9840 & 0.0798 & 0.3829 & 0.7140 & 0.1828 & 0.3635\\
  \midrule
Bagging + Adv. Train. & 0.9927$^*$  & 0.8703$^*$  & 0.9840$^*$  & 0.8320 & 
0.5010$^*$  &  0.7017 \\
Bagging + Def. Dist. & 0.9875 & 0.0954 & 0.4514 & 0.7323 & 0.1839 & 0.4569\\
  \bottomrule
\end{tabular}
\label{compare}
\end{table*}

\section{Conclusion}\label{Conclusion}

With the rise of deep learning as the state-of-the-art approach for many 
classification tasks, researchers noted that neural networks are highly 
vulnerable to adversarial perturbations. This is particularly problematic when 
neural networks are used in security sensitive applications such as autonomous 
driving. Hence, with the development of more efficient attack methods against 
neural networks it is desirable to obtain neural networks that are themselves 
robust against adversarial attacks while showing state of the art performance
on unperturbed data.

In this manuscript, it is shown that several ensemble methods such as  random 
initialization or Bagging do not only increase the accuracy on the test data, 
but also make the classifiers considerably more robust against certain 
adversarial 
attacks. 
We consider ensemble methods as sole defense methods, but more 
robust classifiers can be obtained by combining ensemble methods with other 
defense mechanisms such as adversarial training. However, this typically leads
to a decrease of accuracy on unperturbed data. 
Although only having tested simple attack scenarios, it can be expected that 
ensemble 
methods may improve the robustness against other adversarial attacks.


\end{document}